\def\year{2019}\relax
\documentclass[letterpaper]{article} 
\usepackage{aaai20}  
\usepackage{times}  
\usepackage{helvet} 
\usepackage{courier}  
\usepackage[hyphens]{url}  
\usepackage{graphicx} 
\usepackage{graphicx}  
\usepackage{amsmath}
\usepackage{amssymb}
\usepackage{hyperref}
\usepackage{booktabs}
\usepackage{bm}
\usepackage{textcomp} 
\newcommand{\citet}[1]{\citeauthor{#1}~(\citeyear{#1})}
\usepackage{appendix}
\usepackage{xcolor}
\usepackage{colortbl}
\definecolor{Gray}{gray}{0.9}

\title{Interpretable Outcome Prediction with Sparse \\Bayesian Neural Networks in Intensive Care}

    \author{Hiske Overweg\textsuperscript{\rm 1}\thanks{Equal contribution. Performed during the authors' involvement in the Microsoft AI Residency Program.},
    Anna-Lena Popkes\textsuperscript{\rm 1}$^*$, Ari Ercole\textsuperscript{\rm 2},
    Yingzhen Li\textsuperscript{\rm 1}\thanks{Equal contribution from senior authors},\\
    \Large \textbf{Jos\'{e} Miguel Hern\'{a}ndez-Lobato\textsuperscript{\rm 1,\rm 2,\rm 3}}$^\dagger$, \textbf{Yordan Zaykov\textsuperscript{\rm 1}}$^\dagger$, \textbf{Cheng Zhang\textsuperscript{\rm 1}}$^\dagger$\thanks{Corresponding author: Cheng.Zhang@microsoft.com}\\ 
\textsuperscript{\rm 1}Microsoft Research Cambridge, UK, \textsuperscript{\rm 2} University of Cambridge, UK, \textsuperscript{\rm 3} Alan Turing Institute, UK} 
 
\begin{document}

\maketitle

\begin{abstract}
Clinical decision making is challenging because of pathological complexity, as well as large amounts of heterogeneous data generated as part of routine clinical care. In recent years, machine learning tools have been developed to aid this process. Intensive care unit (ICU) admissions represent the most data dense and time-critical patient care episodes. In this context, prediction models may help clinicians determine which patients are most at risk and prioritize care. However, flexible tools such as artificial neural networks (ANNs) suffer from a lack of interpretability limiting their acceptability to clinicians. In this work, we propose a novel interpretable Bayesian neural network architecture which offers both the flexibility of ANNs and interpretability in terms of feature selection. In particular, we employ a sparsity inducing prior distribution in a tied manner to learn which features are important for outcome prediction.
We evaluate our approach on the task of mortality prediction using two real-world ICU cohorts. In collaboration with clinicians we found that, in addition to the predicted outcome results, our approach can provide novel insights into the importance of different clinical measurements. This suggests that our model can support medical experts in their decision making process.

\end{abstract}

\section{Introduction}

Clinicians often need to make critical decisions, for example, about treatments or patient scheduling, based on available data and personal expertise.  The increasing prevalence of electronic health records (EHRs) means that routinely collected medical data are increasingly available in machine readable form. In some areas such as the intensive care unit (ICU), the data density may be so large that it becomes difficult for clinicians to fully appreciate relationships and patterns in clinical records. At the same time, ICU patients are the most severely ill. Life-supporting treatments are not only expensive and limited but may be associated with potentially catastrophic side effects. It is in this context that appropriate treatments must be delivered in a time-critical manner based on accurate appraisal of all available information. A degree of automated data analysis may assist clinicians navigate the current `data deluge' and make the best informed decisions. The recent success of artificial intelligence and machine learning in various real-world applications suggests that such technology can be the key to unlocking the full potential of medical data and help with making real-time decisions \cite{shipp2002,caruana2015intelligible,bouton2016,chen2017}.

The high data density of the ICU is ideal for applying machine learning methods to assist with clinical decision making. All medical decision making is predicated on the prediction of future outcomes. Especially relevant to the ICU are predictions surrounding patient mortality. As a consequence, several machine learning studies published over the course of the last years focused on this task \cite{joshi2012,celi2012,ghassemi2014,meiring2018}.
 Most of these studies concentrated on improving previously published measures of performance such as discrimination, specificity and sensitivity.

Deploying machine learning solutions in ICUs to support life/death decision making is challenging and requires high prediction accuracy. Artificial neural networks (ANNs) are powerful machine learning models that have been successful in several highly complex real-word tasks \cite{collobert2011natural,boulanger2012,bojarski2016,silver2016}. 
The non-linearity of ANNs allows them to capture complex non-linear dependencies, a quality which often results in high predictive performance. Despite widespread success, predictions from ANNs lack interpretability. Instead, they often function as a black box. For example, after training an ANN on the task of outcome prediction it is difficult to determine which input features are relevant for making predictions. This is highly undesirable in the medical domain --- making potentially life-changing decisions without being able to clearly justify them is unacceptable to both clinicians and patients. As a consequence, the application of ANNs in practice has been limited. Advancing the interpretability of such networks is 
essential to increase their impact in healthcare applications.

In this work, we propose an interpretable machine learning model based on a Bayesian neural network (BNN) \cite{mackay:practical1992,hinton:mdl1993,blundell2015,hernandez-lobato:pbp2015,louizos2017,ghosh2017} for outcome prediction in the ICU. Our proposed method offers not only the flexibility of ANNs but also interpretable predictions 
--- inspecting the model parameters directly shows which features are 
considered irrelevant for prediction. In particular, we propose the use of tied sparsity inducing prior distributions, where the same sparsity prior is shared among all weights connected to the same input feature. For some of the input features, all of their connecting weights will be suppressed after training, indicating that the corresponding input features are not relevant for  prediction. For the prior distribution we use the horseshoe prior because of its sparsity inducing and heavy-tailed nature. 
Furthermore a BNN explicitly models the uncertainty in a dataset, as well as in the model parameters and predictions. This replicates the inherently probabilistic nature of clinical decision making. 



 
We apply our method to two real-world ICU cohorts, MIMIC-III \cite{johnson2016} and CENTER-TBI \cite{Maas2014}, to predict patient outcome. The model provides
fully probabilistic predictions.
Such results can help clinicians with making treatment decisions and communicating with patients' families. 
More importantly, our method
allows to determine which medical measures are irrelevant for the task of outcome prediction. This is an important advance because model interpretability is essential if critical decisions based on diagnostic support systems are to be accepted by both clinicians and the public.\\

\section{Related work}
\label{sec:related}
\paragraph{Sparsity in linear models}
In linear models sparsity is typically induced using a suitable prior distribution over the model parameters. The well known LASSO (Least Absolute Shrinkage and Selection Operator) \cite{tibshirani1996} produces a sparse estimate of the parameters in a linear model, by regularizing the $\ell_1$ norm of the parameter vector. 
\citet{park2008} showed that the $\ell_1$ regularizer can be interpreted as a Laplace prior over the parameters, and that the LASSO estimate is equivalent to a maximum a posteriori estimate of the linear coefficients given data.

Sparsity inducing prior distributions beyond the Laplace distribution have been proposed to improve feature selection results.
For example, the spike and slab prior places a mixture distribution on each parameter, comprising a point mass distribution at zero (the spike) and an absolutely continuous density (the slab) \cite{mitchell}. 

The spike and slab prior requires a careful choice of the mixture weights and the variance of the ``slab''. 
Another popular choice for introducing sparsity is the horseshoe prior \cite{carvalho}, which assigns a half-Cauchy prior over the variance of the Gaussian prior over the parameters. The heavy tail of the horseshoe distribution allows coefficients associated with important features to remain large, while at the same time the tall spike at the origin encourages shrinkage of other parameters. Compared to the spike and slab prior, the horseshoe prior is more stable and more computationally efficient.
Therefore, in
this work we employ the horseshoe prior as the sparsity inducing prior for 
feature selection.

\paragraph{Sparsity in non-linear models} 

Less work has been directed towards the application of sparsity inducing priors in non-linear models. One of the first approaches was Automatic Relevance Determination applied to BNNs \cite{mackay1994}, which fits the prior variance for each individual parameters by maximizing the marginal likelihood. However, this approach fails to scale to large datasets as it involves the inversion of large matrices.
\citet{louizos2017} and \citet{ghosh2017} applied a horseshoe prior to prune inactive hidden units from BNNs, thereby achieving better compression.
Our work uses the same inference techniques, however, we focus on selecting input features, not hidden units.

\paragraph{Outcome prediction in the medical domain}

Because of its central importance to patients and clinicians, outcome prediction for ICU patients is a widely studied task. Models can be divided into two categories: those using only static features \cite{knaus1985,le1993,lemeshow1993,elixhauser1998,steyerberg2008} and those utilizing information about the temporal evolution of features \cite{joshi2012,ghassemi2014,caballero2015,harutyunyan2017,che2018}.

Most approaches based on static features model only \textit{linear} relationships or rely on manual feature engineering. Manual feature engineering scales poorly, and prevents models from automatically discovering patterns in the data. Linear models are easy to interpret, because the importance of input features can directly be inferred from the magnitude of the associated model coefficients. This is highly desirable for transparent clinical decision making, but
the capacity of linear models is limited. In most real world problems the relationship between input features and target values is \textit{non-linear} or may involve complex interactions between predictors. Consequently, more powerful approaches are required.

In this work, we propose a model for mortality prediction named HorseshoeBNN. In contrast to previous work \cite{joshi2012,celi2012,ghassemi2014,caballero2015} our model is able to both capture non-linear relationships and learn which input features are important for prediction.

\section{Methods}
\label{sec:method}

In this section, we describe our proposed BNN architecture for mortality prediction. We first revisit the BNN, a type of ANN which explicitly models uncertainty by introducing distributions over the model parameters. We then introduce the specific prior distribution we employ, the horseshoe prior, which induces sparsity in the first layer of the BNN, thereby enabling features selection. Finally we describe the architecture of our model, the HorseshoeBNN, and discuss the computational methods to implement it. 

\subsection{Bayesian Neural Networks}
\label{sec:bnn}

Given an observed dataset $\mathcal{D} = \{(\bm{x}_n, \bm{y}_n) \}_{n=1}^N$, an ANN uses a set of parameters $\bm{\theta} $ to determine a model $\bm{y} = f(\bm{x}; \bm{\theta})$ that fits the data well and generalizes to unseen cases. Instead of directly predicting the response $\bm{y}$ with a deterministic function $f$, BNNs start from a probabilistic description of the modelling task, and estimate the uncertainty of the parameters given the data. Concretely, the network parameters $\bm{\theta}$ are considered random variables, and a prior distribution $p(\bm{\theta})$ is selected to represent the prior belief of their configuration. 
Assuming that the observed data is independent and identically distributed (i.i.d.), the \emph{likelihood function} of $\bm{\theta}$ is defined as 
\begin{equation}
    p(\mathcal{D} | \bm{\theta}) = \prod_{n=1}^N p(\bm{y}_n | \bm{x}_n, \bm{\theta}), 
\end{equation}
where, in case of a binary classification task like the one presented in this work, the label $y_n$ is a scalar, and
\begin{equation}
\begin{split}
    \log p(y_n | \bm{x}_n, \bm{\theta}) =~& y_n \log(f(\bm{x}_n); \bm{\theta}) \\ &+ (1 - y_n) \log(1-f(\bm{x}_n; \bm{\theta})).
\end{split}
\end{equation}
For regression tasks, we have 
$
p(\bm{y}_n | \bm{x}_n, \bm{\theta}) = \mathcal{N}(\bm{y}_n; f(\bm{x}_n; \bm{\theta}), \sigma^2 \mathbf{I}).
$
After observing the training data $\mathcal{D}$, a \emph{posterior distribution} of the network weights $\bm{\theta}$ is defined by Bayes' rule
\begin{equation}
    p(\bm{\theta} | \mathcal{D}) = \frac{p(\bm{\theta}) p(\mathcal{D} | \bm{\theta})}{p(\mathcal{D})}, \quad p(\mathcal{D}) = \int p(\bm{\theta}) p(\mathcal{D} | \bm{\theta}) d\bm{\theta}.
\end{equation}
This posterior distribution represents the \emph{updated} belief of how likely the network parameters are given the observations. 
It can be used to predict the response $\bm{y}^*$ of an unseen input $\bm{x}^*$ using the \emph{predictive distribution}:
\begin{equation}
    p(\bm{y}^* | \bm{x}^*, \mathcal{D}) = \int p(\bm{y}^* | \bm{x}^*, \bm{\theta}) p(\bm{\theta} | \mathcal{D}) d\bm{\theta}.
\end{equation}

\subsection{The HorseshoeBNN: Feature Selection with Sparsity Inducing Priors}
\label{ss:horseshoe_bnn}

\begin{figure}[h!]
\centering
         \includegraphics[width= 0.25 \textwidth]{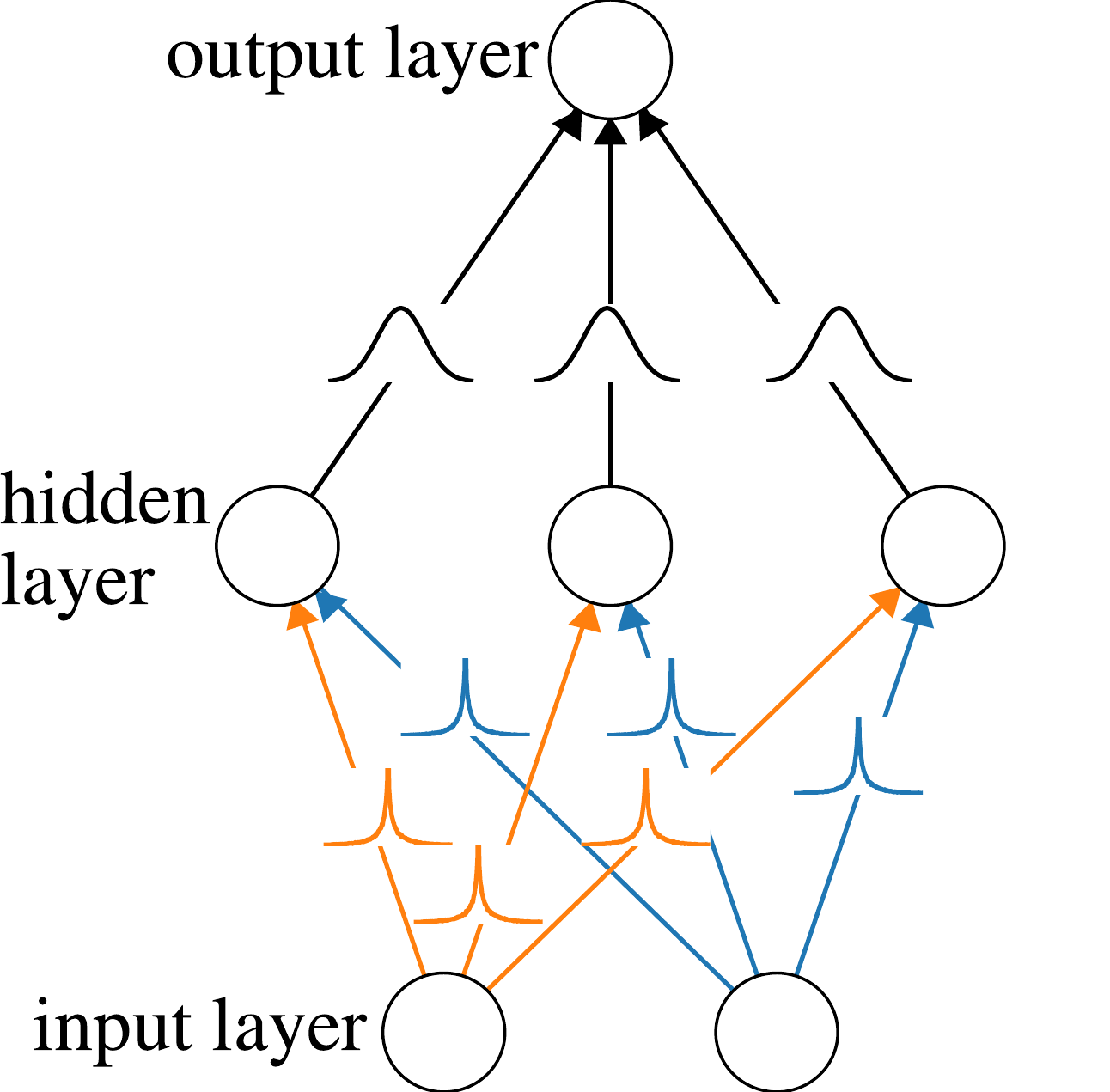}

    \caption{ Illustration of a HorseshoeBNN. The prior distribution of the weights in the input layer is given by a horseshoe distribution. All weights for particular input feature share the same shrinkage parameter, allowing for feature selection. The prior of the weights of the second layer is given by a Gaussian.
    \vspace{-10pt}
    }
    \label{fig:bnn}
\end{figure}

The prior distribution $p(\bm{\theta})$ captures the prior belief about which model parameters are likely to generate the target outputs $\bm{y}$, before observing any data. When focusing on feature selection, sparsity inducing priors are of particular interest. In this work, we use a horseshoe prior \cite{carvalho}, which 
can be described as
\begin{equation}
w|\tau \sim \mathcal{N}(0, \tau^2) \quad\mathrm{where}\quad \tau \sim C^+(0, b_0),
\label{eq:horseshoe}
\end{equation}
where $C^+$ is the half-Cauchy distribution and $\tau$ is a scale parameter. The probability density function of the horseshoe prior with $b_0=1$ is illustrated in Figure \ref{fig:bnn}. It has a sharp peak around zero and wide tails. This encourages shrinkage of weights that do not contribute to prediction, while at the same time allowing large weights to remain large. 

For feature selection we propose a horseshoe prior for the first layer of a BNN by using a shared half-Cauchy distribution to control the shrinkage of weights connected to the same input feature. Specifically, denoting $W_{ij}^{(1)}$ as the weight connecting the $j$-th component of the input vector $\bm{x}$ to the $i$-th node in the first hidden layer, the associated horseshoe prior is given by

\begin{equation}
W_{ij}^{(1)} | \tau_n, v \sim \mathcal{N}(0, \tau_j^2 v^2)
\end{equation}
where $\tau_j \sim C^+(0,b_0)$ and $v \sim C^+(0,b_g)$.

\noindent
The layer-wide scale $v$ tends to shrink all weights in a layer, whereas the local shrinkage parameter $\tau_j$ allows for reduced shrinkage of all weights related to a specific input feature $x_j$. As a consequence, certain features of the input vector $\bm{x}$ are selected whereas others are ignored. For the bias node we use a Gaussian prior distribution. It is important to note that a simpler solution which keeps all feature weights small but does not encourage significant shrinkage would not be sufficient 
for feature selection --- large weights in deeper layers of the network might increase the influence of irrelevant input features.

We model the prior of the weights in the second layer of the HorseshoeBNN by a Gaussian distribution, which prevents overfitting \cite{blundell2015}. The complete network architecture is given in Figure \ref{fig:bnn}. Although we perform our experiments with a single hidden layer, the model can easily be enlarged by adding more hidden layers with Gaussian priors. 

A direct parameterization of the half-Cauchy prior can lead to instabilities in variational inference (VI) for BNNs. Thus, we follow \citet{ghosh2017} to reparametrize the horseshoe prior using auxiliary parameters:
\begin{equation}
\begin{split}
&a \sim C^+(0, b) \quad \Leftrightarrow \\& a|\kappa \sim \text{Inv} \, \Gamma (\frac{1}{2}, \frac{1}{\kappa});\quad \kappa \sim \text{Inv} \, \Gamma(\frac{1}{2}, \frac{1}{b^2}).
\end{split}
\end{equation}

\noindent After adding the auxiliary variables to the Horseshoe prior, the prior over all the unobserved random variables $\bm{\theta} = \{ \{ \bm{W}^{(l)} \}_{l=1}^{L+1}, v, \vartheta, \bm{\tau} = \{\tau_j \}, \bm{\lambda} = \{ \lambda_j \} \}$ is
%
$$
p(\bm{\theta}) = p(\bm{W}^{(1)}, v, \vartheta, \bm{\tau}, \bm{\lambda}) \prod_{l=2}^{(L+1)} p (\bm{W}^{(l)}),
$$
\vspace{-5pt}
$$
p(\bm{W}^{(l)})= \prod_{i,j} \mathcal{N}(W_{ij}^{(l)}; 0, \sigma^2), \quad l = 2, ..., L+1,
$$
\begin{equation}
\begin{split}
&p(\bm{W}^{(1)}, v, \vartheta, \bm{\tau}, \bm{\lambda}) = 
\\&p(v | \vartheta) \, p(\vartheta) \prod_{j} p(\tau_j | \lambda_j) \, p(\lambda_j) \prod_{i} \mathcal{N}(W_{ij}^{(1)}; 0, \tau_{j}^2 v^2), \nonumber
\end{split}
\end{equation}
\begin{equation}
\begin{split}
& p(\tau_j | \lambda_j) = \text{Inv} \, \Gamma (\frac{1}{2}, \frac{1}{\lambda_j}), \quad p(\lambda_j) = \text{Inv} \,  \Gamma (\frac{1}{2}, \frac{1}{b_0^2}), 
\end{split}
\end{equation}
\begin{equation}
\begin{split}
p(v | \vartheta) = \text{Inv} \, \Gamma (\frac{1}{2}, \frac{1}{\vartheta}), \quad p(\vartheta) = \text{Inv} \, \Gamma (\frac{1}{2}, \frac{1}{b_g^2}) \nonumber.
\end{split}
\end{equation}

\subsection{Scalable Variational Inference for HorseshoeBNN}
\label{ss:implementation}
For most BNN architectures both the posterior distribution $p(\bm{\theta} | \mathcal{D})$ and the predictive distribution $p(\bm{y}^* | \bm{x}^*, \mathcal{D})$ are intractable due to a lack of analytic forms for the integrals. To address this outstanding issue we fit a simpler distribution $q_{\phi}(\bm{\theta}) \approx p(\bm{\theta} | \mathcal{D})$ and later replace $p(\bm{\theta} | \mathcal{D})$ with $q_{\phi}(\bm{\theta})$ in prediction.
More specifically, we define
\begin{equation}
\begin{split}
    q_{\phi}(\bm{\theta} ) =~&q_{\phi}(\bm{W}^{(1)}| \bm{\tau}, v) \, q_{\phi}(v) \, q_{\phi}(\vartheta) \,  q_{\phi}(\bm{\tau}) \, q_{\phi}(\bm{\lambda})\\ 
    &\times \prod_{l=2}^{L+1} q_{\phi}(\bm{W}^{(l)}),
\end{split}
\end{equation}
and use factorized Gaussian distributions for the weights in upper layers: 
$$q(\bm{W}^{(l)}) = \prod_{i,j} \mathcal{N}(W_{ij}^{(l)}|\mu_{W_{ij}^{(l)}}, \sigma^2_{W_{ij}^{(l)}}), \quad l = 2, ..., L+1.$$
To ensure non-negativity of the shrinkage parameters, we consider a log-normal approximation to the posterior of $v$ and $\tau_j$, i.e.~
\begin{equation}
    q_{\phi}(v) = \mathcal{N}(\log v; \mu_v, \sigma^2_v), \quad q_{\phi}(\tau_j) = \mathcal{N}(\log \tau_j; \mu_{\tau_j}, \sigma^2_{\tau_j}).
\label{eq:logn}
\end{equation}

\noindent
In the horseshoe prior (see Eq. \ref{eq:horseshoe}) the weights $W_{ij}$ and the scales $\tau_{l}$ and $v$ are strongly correlated. This leads to strong correlations in the posterior distribution with pathological geometries that are hard to approximate. \citet{Betancourt2015} and \citet{Ingraham2016} show that this problem can be mitigated by reparametrizing the weights in the horseshoe layer as follows:
\begin{equation}
\beta_{ij} \sim \mathcal{N}(\beta_{ij}|\mu_{\beta_{ij}}, \sigma^2_\beta{_{ij}}), \quad W_{ij}^{(1)} = \tau_l v \beta_{ij},
\end{equation}
and equivalently, parametrizing the approximate distribution $q(\bm{W}^{(1)} | v, \bm{\tau})$ as 

\begin{equation}
\begin{split}
  q(\bm{W}^{(1)} | v, \bm{\tau}) = &\prod_{i,j}  q(W_{ij}^{(1)} | v, \tau_j) \\= &\prod_{i,j} \mathcal{N}(W_{ij}^{(1)}; v \tau_j \mu_{\beta_{ij}}, v^2 \tau_j^2 \sigma^2_\beta{_{ij}}).
\end{split}
\end{equation}
Because the log-likelihood term $p(\bm{y} | \bm{x}, \bm{\theta})$ does not depend on $\vartheta$ or $\bm{\lambda}$, one can show that the optimal approximations $q(\vartheta)$ and $q(\bm{\lambda})$ are inverse Gamma distributions with distributional parameters dependent on $q(\bm{\theta} \backslash \{ \vartheta, \bm{\lambda} \} )$ \cite{ghosh2017}. 

We fit the variational posterior $q_{\phi}(\bm{\theta})$ by minimizing the Kullback-Leibler (KL) divergence $\mathrm{KL}\big[q_{\phi}(\bm{\theta}) || p(\bm{\theta} | \mathcal{D}) \big]$. One can show that the KL divergence minimization task is equivalent to maximizing the \emph{evidence lower-bound} (ELBO) \cite{jordan:vi1999,beal:vi2003,zhang2018advances}
\begin{equation}
\begin{split}
    \mathcal{L}(\phi) = & \mathbb{E}_{q_{\phi}(\bm{\theta} )} \big[ \log p(\mathcal{D} | \bm{\theta}) \big] - \mathrm{KL}\big[q_{\phi}(\bm{\theta} ) || p(\bm{\theta})\big] \\ = & \mathbb{E}_{q_{\phi}(\bm{\theta} )} \big[\log p(\mathcal{D})\big] - \mathrm{KL}\big[q(\bm{\theta} | \phi) || p(\bm{\theta} | \mathcal{D})\big]. \nonumber
\end{split}
\end{equation}
Since the ELBO still lacks an analytic form due to the non-linearity of the BNN, we apply black box VI \cite{ranganath:bbvi2014} to compute an unbiased estimate of the ELBO by sampling $\bm{\theta} \sim q_{\phi}(\bm{\theta})$. More specifically, because the $q$ distribution is constructed by a product of (log-) normal distributions, we apply the reparametrization trick \cite{kingma:vae2014,rezende:vae2014} to draw samples from the variational distribution:
$
    w \sim \mathcal{N}(w; \mu, \sigma^2) \Leftrightarrow \epsilon \sim \mathcal{N}(\epsilon; 0, 1), w = \mu + \sigma \epsilon.
$
Furthermore, stochastic optimization techniques are employed to allow for mini-batch training, which enables the VI algorithm to scale to large datasets. Combining both, the \emph{doubly stochastic} approximation to the ELBO is
\begin{equation}
\begin{split}
    &\mathcal{L}(\phi) \approx \frac{N}{M} \sum_{m=1}^M \log p(\bm{y}_m | \bm{x}_m, \bm{\theta}) - \mathrm{KL}\big[q_{\phi}(\bm{\theta}) || p(\bm{\theta})\big], \\&\bm{\theta} \sim q_{\phi}(\bm{\theta}), \ \{ (\bm{x}_m, \bm{y}_m) \}_{m=1}^M \sim \mathcal{D}^M, \nonumber
\end{split}
\end{equation}
which is used as the loss function for the stochastic gradient ascent training of the variational parameters $\phi$.

\section{Experiments and Results}
\label{sec:exp}

We evaluate our model on several datasets. In Appendix \ref{app:sparsity} we present a validation study on a synthetic feature selection dataset, showing that our approach can recover the ground truth set of features. Here we present a performance benchmark on datasets from the UCI repository and then evaluate our method on two real-world ICU cohorts: MIMIC-III \cite{johnson2016} and CENTER-TBI \cite{Maas2014}.
\subsection{Experimental Set-Up}
\label{sec:setup}

We compare our proposed HorseshoeBNN with the following methods. 

\begin{itemize}
\setlength\itemsep{-0.2em}
    \item \textit{LinearGaussian}: a linear model with a Gaussian prior distribution on all weights \cite{bishop2006}. This is the most commonly used Bayesian model for prediction tasks.
    \item \textit{GaussianBNN}: a standard BNN with a Gaussian prior distribution on all weights \cite{blundell2015}. 
    \item \textit{LinearHorseshoe}: a linear model with a horseshoe prior distribution on all weights \cite{carvalho}. This model uses sparsity inducing prior distributions for performing feature selection.
    \item \textit{HorseshoeBNN}: our novel extension of the GaussianBNN with a tied horseshoe prior distribution on the weights in the first layer. This enables the model to perform feature selection.
    \item \textit{Lasso}: For details, see Appendix \ref{app:models}.
    \item \textit{SupportVectorMachine} (SVM): see Appendix \ref{app:models}.
    \item \textit{RandomForest}: see Appendix \ref{app:models}.
\end{itemize}

\noindent All models are trained until convergence using 10 fold cross-validation and the ADAM optimizer \cite{kingma2014}. We use 50 hidden units for the UCI datasets and MIMIC-III cohort and 100 hidden units for CENTER-TBI. The full list of hyperparameter settings can be found in Appendices \ref{app:mimic} and \ref{app:tbi}.\footnote{The code for the experiments is available at \url{https://github.com/microsoft/horseshoe-bnn}}

\subsection{UCI benchmark}

We report the root mean squared error (RMSE) and negative log-likelihood (NLL) results on three regression datasets from the UCI repository. The GaussianBNN and the Horsesehoe BNN perform better than the linear models. The RandomForest performs best in terms of RMSE on these small datasets. However, it does not provide a principled probabilistic estimation and its performance is worse for the clinical datasets discussed below.
\begin{table}[h!]
    \small
\begin{center}
\scalebox{0.8}{
    \begin{tabular}{c  c  c  c }
    \toprule
    \textbf{Dataset}&
    \textbf{Model} &
    \textbf{RMSE} &
     \textbf{NLL}  \\
    \midrule
    Boston&LinearGaussian & 4.806 $\pm$ 0.191 &  2.992 $\pm$ 0.039\\
    &GaussianBNN & 3.470 $\pm$ 0.283 & 2.646 $\pm$ 0.085 \\
    &LinearHorseshoe & 4.840 $\pm$ 0.206 & 3.018 $\pm$ 0.038 \\
    &HorseshoeBNN &  3.500 $\pm$ 0.243 &  \textbf{2.617} $\pm$ 0.062 \\
    &Lasso & 4.79 $\pm$ 0.19 & -  \\
    &SVM& 4.990$\pm$ 0.254 & -\\
    &RandomForest &  \textbf{3.151}$\pm$ 0.145& - \\
     \midrule
    Yacht &LinearGaussian &  9.008 $\pm$ 0.366 & 3.627 $\pm$ 0.039\\
    &GaussianBNN & 1.157 $\pm$ 0.069 & 1.582 $\pm$ 0.030 \\
    &LinearHorseshoe & 8.921 $\pm$ 0.367 & 3.616 $\pm$ 0.037 \\
    &HorseshoeBNN &\textbf{0.959} $\pm$ 0.091 & \textbf{1.334} $\pm$ 0.139 \\
    &Lasso & 8.95 $\pm$ 0.45 & -  \\
    &SVM& 10.715$\pm$ 0.831 & -\\
    &RandomForest & 0.989$\pm$ 0.110 & - \\
     \midrule
    Wine &LinearGaussian & 0.652 $\pm$ 0.013 &  0.993 $\pm$ 0.020 \\
    &GaussianBNN &  0.640 $\pm$ 0.014 &   \textbf{0.974} $\pm$ 0.019 \\
    &LinearHorseshoe & 0.635 $\pm$ 0.013 & 0.966 $\pm$ 0.020 \\
    &HorseshoeBNN & 0.639 $\pm$ 0.012 &   \textbf{0.974} $\pm$ 0.016\\
    &Lasso & 0.652 $\pm$ 0.013 &  -  \\
    &SVM & 0.654$\pm$ 0.015 & -\\
    &RandomForest & \textbf{0.570}$\pm$ 0.010 & - \\
        \bottomrule
    \end{tabular}
    }
\end{center}
\caption{Test performance in RMSE and negative log likelihood, the lower the better.
\vspace{-10pt}
}
\label{tab:uci}
\end{table}
\subsection{Mortality prediction on MIMIC-III} 
\label{sub:mimic-results}
\paragraph{Cohort}
MIMIC-III (‘Medical Information Mart for Intensive Care’) \cite{johnson2016} is a 
publicly available intensive care database collected from tertiary care hospitals.
The data includes information about laboratory measurements, medications, notes from care providers and other features. 
In total, the database contains medical records of 7.5K ICU patients over 11 years.

\paragraph{Preprocessing}
We preprocess MIMIC-III using code introduced in \citet{harutyunyan2017}, focusing on the task of mortality prediction. For all measurements we remove invalid feature values beyond the allowed range 
defined by medical experts (see
Table \ref{tab:mimic} in Appendix \ref{app:mimic}). 
Since we do not focus on prediction based on dynamic features, we reduce the time series for each patient by computing the mean value of each feature in the first 48 hours of the patient's stay in the ICU. We use mean imputation for missing values. The final cohort contains 17903 samples and 17 features. 86.5\% of patients contained in the dataset survived, 13.5\% deceased. All features are listed in Table \ref{tab:mimic} in Appendix \ref{app:mimic}.

\paragraph{Prediction}

We present the error rate, the area under the receiver operating characteristic curve (AUROC) and negative predictive log-likelihood results in Table \ref{tab:metrics_mimic}. 
Overall, the BNNs perform better than the other models. The HorseshoeBNN performs on par with or slightly better than the GaussianBNN, potentially because better feature selection reduces overfitting. We also present the confusion matrix for the Horseshoe models in Figure \ref{fig:mimic_confusion}. Although the dataset is imbalanced with a significantly smaller amount of deceased patient data, the HorseshoeBNN still show an improvement in correctly predicting the outcome of deceased patients compared to the LinearHorseshoe model.

\begin{table*}[t]
\begin{minipage}[t]{0.59\textwidth}
    \small
\begin{center}
\vspace{-90pt}
\scalebox{0.9}{
    \begin{tabular}{  c  c  c  c }
    \toprule
    \textbf{Model} &
    \textbf{Error rate} &
    \textbf{AUROC} & \textbf{NLL}  \\
    \midrule
    LinearGaussian & 0.129 $\pm$ 0.003 & 0.807 $\pm$ 0.004 & 0.321 $\pm$ 0.004\\
    GaussianBNN & 0.123 $\pm$ 0.003  & 0.830 $\pm$ 0.004 & \textbf{0.304} $\pm$ 0.004 \\
    LinearHorseshoe & 0.130 $\pm$ 0.003  & 0.807 $\pm$ 0.004 & 0.320 $\pm$ 0.004 \\
    HorseshoeBNN & \textbf{0.122} $\pm$ 0.002 & \textbf{0.831} $\pm$ 0.004 & \textbf{0.304} $\pm$ 0.004 \\
    Lasso & 0.129 $\pm$ 0.002 & 0.795 $\pm$ 0.004 & 0.325 $\pm$ 0.004\\
    SVM & 0.129 $\pm$ 0.003 & - &  -\\
    RandomForest & 0.125$\pm$ 0.002 & - & - \\
    \bottomrule
    \end{tabular}
    }
\end{center}
\caption{Results of the different models for the task of mortality prediction tested on the MIMIC-III cohort. NLL is the negitive log-likelihood. The mean value and standard error of each metric over 10-fold cross-validation is presented.}
\label{tab:metrics_mimic}
\end{minipage}
\hfill
\begin{minipage}[t]{0.37\textwidth}
    \centering
    \includegraphics[width = 0.95 \textwidth]{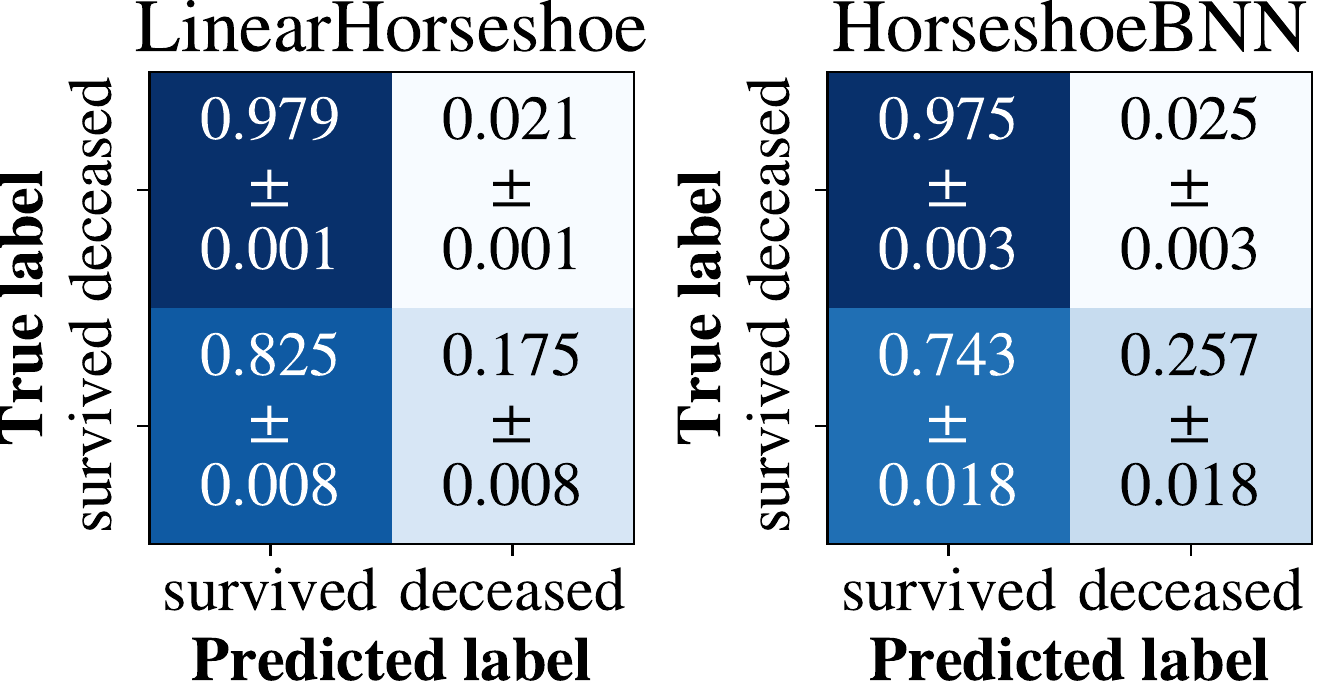}
    \caption{Confusion matrices of the horseshoe models trained on the MIMIC-III cohort (10-fold cross-validated).}
    \label{fig:mimic_confusion}
\end{minipage}
\end{table*}

\paragraph{Interpretability and Clinical Relevance}

\begin{figure*}[h!]
    \centering
    \includegraphics[width = 0.89\textwidth]{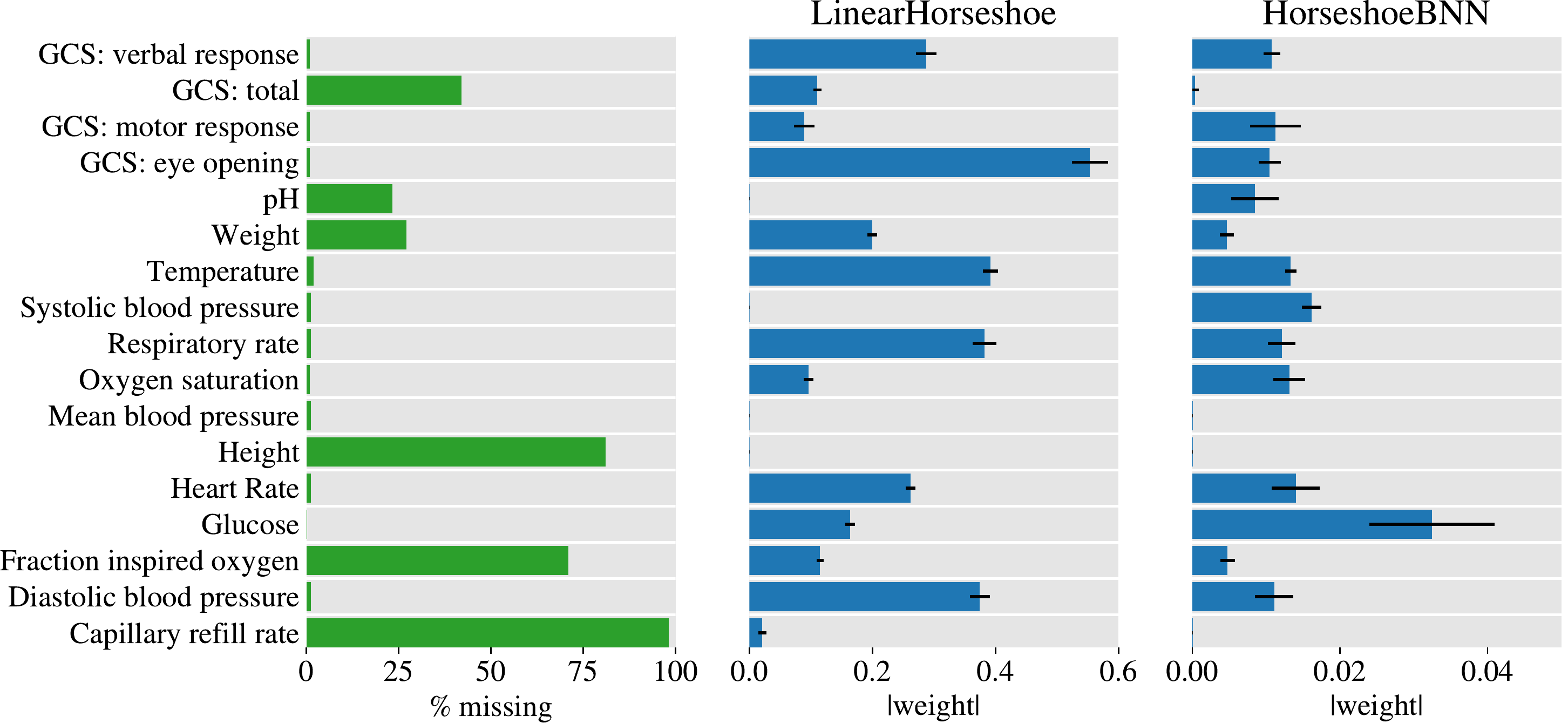}
    \caption{\textbf{Left:} percentage of missing data for the features in the MIMIC-III cohort. \textbf{Middle:} Norm of the weights of the LinearHorseshoe model, representing the relative importance of the corresponding features. \textbf{Right:} Norm of the weights of the HorseshoeBNN. The name of each input feature is given on the far left of the plots. Feature weights of zero indicate that the corresponding features are irrelevant for outcome prediction. All non-zero weights indicate that the corresponding features are relevant for predicting mortality.
    \vspace{-5pt}
    }
    \label{fig:importance_mimic}
\end{figure*}

In addition to improved predictive performance, the HorseshoeBNN allows for better feature selection. For linear models we inspect the posterior mean of the weights directly connected to the features. For BNNs we look at the average of the posterior means of the outgoing weights. We plot a histogram of these weight values with logarithmic bins (see Appendix \ref{app:mimic}). For the horseshoe models, the histogram shows two groups of weights which differ by orders of magnitude, facilitating the choice of a threshold for feature relevance. We verify that the metric values remain unchanged when training a horseshoeBNN without the features considered irrelevant by the model. The clear dichotomy in the histogram of the weights is absent for Gaussian models (see Appendix \ref{app:mimic}), showing the importance of the horseshoe prior to efficiently eliminate irrelevant features.

The ability to perform feature selection also improves the interpretability of the model predictions. The weight values described above reflect the relevance of the features and
  are  visualised  in  Figure \ref{fig:importance_mimic}. We find that the LinearHorseshoe model and the HorseshoeBNN agree on the relevance of most features except for \textit{pH, systolic blood pressure} and \textit{Glasgow coma scale}.

The HorseshoeBNN picks up the pH feature, whereas the LinearHorseshoe model does not, presumably because the outcome depends on pH values in a non-linear way (which a linear model cannot capture): the healthy range for pH is very narrow and both too high and too low values are dangerous.

For blood pressure, the LinearHorseshoe model captures the diastolic blood pressure only, while the HorseshoeBNN captures both the diastolic and systolic blood pressure. Although recording diastolic pressure is sufficient to establish a baseline blood pressure for a patient, combining diastolic and systolic values allows to obtain additional information about the waveform of a patient's blood pressure, which can be of clinical relevance. This suggests that the HorseshoeBNN is able to recognize the importance of this extra information, whereas the LinearHorseshoe model is not.

The feature \textit{Glasgow coma scale total} is selected only by the LinearHorseshoe model, but considered irrelevant by the HorseshoeBNN. 
Note the Glasgow coma total scale is the sum of the verbal response, motor response and eye opening, and the latter three features are all selected by the HorseshoeBNN.

Two features, namely \textit{height} and \textit{capillary refill rate (CRR)} are considered irrelevant by both models. 
While height is considered not informative by the domain expert, CRR is expected to be relevant. However, CRR is observed for an extremely small fraction of patients, which can explain why the model did not consider it.

Overall these results suggest that, compared to a linear model, a non-linear model like our HorseshoeBNN can provide additional insights into feature relevance.

\subsection{Mortality prediction on CENTER-TBI} 
\label{sub:tbi-results}
\paragraph{Cohort}
The CENTER-TBI (Collaborative European NeuroTrauma Effectiveness Research in Traumatic Brain Injury) study is an 
observational study 
conducted across Europe and Israel \cite{Maas2014}. It contains data from 5400 patients with 
traumatic brain injury (TBI). 
We focus on the subset of patients
in the emergency room and admitted to the ICU. The cohort contains a broad
range of clinical data, including baseline demographics, 
mechanism of injury, 
vital signs, Glasgow coma scale, and
brain computed tomographic reports and many other features.

\begin{table}[t]
    \small
\begin{center}
\scalebox{0.9}{
    \begin{tabular}{  c  c  c  c }
    \toprule
    \textbf{Model} &
    \textbf{Error rate} &
    \textbf{AUROC} & \textbf{NLL}  \\
    \midrule
    LinearGaussian & 0.185 $\pm$ 0.008 & 0.871 $\pm$ 0.013 & 0.393 $\pm$ 0.018\\
    GaussianBNN & 0.195 $\pm$ 0.010 & 0.869 $\pm$ 0.013 & 0.390 $\pm$ 0.020\\
    LinearHorseshoe & 0.180 $\pm$ 0.008 & \textbf{0.874} $\pm$ 0.013 & 0.383 $\pm$ 0.016 \\
    HorseshoeBNN & \textbf{0.179} $\pm$ 0.008 & 0.873 $\pm$ 0.013 & \textbf{0.380} $\pm$ 0.014 \\
    Lasso & 0.192 $\pm$ 0.009 & 0.844 $\pm$ 0.013 & 0.423 $\pm$ 0.013 \\
    SVM & 0.182 $\pm$ 0.007& - & - \\
    RandomForest & 0.182$\pm$ 0.011 & - & - \\
    \bottomrule
    \end{tabular}
    }
\end{center}
\caption{Results of the different models for the task of mortality prediction tested on the CENTER-TBI cohort. NLL is the negitive log-likelihood. The mean value and standard error of each metric over 10-fold cross-validation is presented.}
\label{tab:metrics_tbi}
\end{table}
\begin{table}[h!]
    \small
\begin{center}
\scalebox{0.8}{
    \begin{tabular}{ c  c  c  c }
    \toprule
     \textbf{Feature}  &Linear &Horseshoe &IMPACT\\
     &Horseshoe&BNN&\\
     \midrule
     Age & x & x & x\\
     Sex & x & x & \\
     Heart rate & x & x & \\
     pH & x & x & \\
     Hypoxia & x &  & x\\
     Hypotension &  &  &x \\
     GCS: motor response & x & x & x\\
     Pupil reaction & x & x & x\\
     Marshall CT Classification & x & x & x\\
     Subarachnoid hemorrhage & & & x\\
     Absent basal cisterns & x & x & \\
     Extradural Hematoma & x & x & x\\
     Glucose & x & x & x\\
     Hemoglobin & & &x\\
     International normalized ratio &x &x &\\
     23 remaining features & & & \\
    \end{tabular}
    }
\end{center}
\caption{List of features marked as relevant by the models. Features marked with an x are relevant.
\vspace{-5pt}
}
\label{tab:tbi_rel}
\end{table}

\newpage
\paragraph{Preprocessing}
We predict mortality based on the features listed in Table \ref{table:TBI} of appendix \ref{app:tbi}, using release 1.0 of the CENTER-TBI cohort. We remove the data of patients for which no outcome was reported. To address missing values we use zero-imputation for binary features and mean imputation for continuous and ordinal features, as suggested by clinicians. The final cohort contains 1613 samples and 38 features (see Appendix \ref{app:tbi}). 75\% of patients in the cohort survived, 25\% deceased. 

\paragraph{Prediction}
The results 
are summarized in Table \ref{tab:metrics_tbi} and confusion matrices are shown in Appendix \ref{app:tbi}. We again observe that the HorseshoeBNN achieves slightly better performance than the linear models. The GaussianBNN performs worse than all other models. This could be due to the large amount of noise in the data which the GaussianBNN might be modeling, thereby overfitting the data. In contrast, the HorseshoeBNN removes input features which makes the model less likely to overfit.

\paragraph{Interpretability and Clinical Relevance}
The relevance of the input features as determined by the LinearHorseshoe model, the HorseshoeBNN and the RandomForest are shown in Figures \ref{fig:importance_tbi} and \ref{fig:importance_forest} in Appendix \ref{app:tbi}. Most of the weights of the RandomForest  are similar, making this model unsuitable for feature selection. In Table \ref{tab:tbi_rel} we compare the features identified as relevant by the LinearHorseshoe model and the HorseshoeBNN. We also list features included in the IMPACT model \cite{steyerberg2008}, a model commonly used for outcome prediction in TBI. We observe a large overlap between features used in the IMPACT model and features considered important by our horseshoe models, which is an indication that our feature selection method works correctly. The LinearHorseshoe model and the HorseshoeBNN attribute similar relevance to most features.

Interestingly, our models attribute little relevance to events of \textit{hypotension}, although such events are considered important from a clinical perspective. A clinically likely explanation for this is that hypotension is typically unrecognised, leading to many false negatives for this feature.

Similar to hypotension, both models consider the feature \textit{hemoglobin} irrelevant for mortality prediction. This might be explained by the small role of hemoglobin in the IMPACT model. It is also possible that the small effect of hemoglobin is already accounted for by the feature \textit{international normalized ratio}. Furthermore, both models select 
certain features 
that are not part of the IMPACT model: \textit{sex, heart rate, pH, international normalized ratio} and \textit{absent basal cisterns}.
In Appendix \ref{app:feature-selection-results} we compare the feature relevance determined for the CENTER-TBI and the MIMIC-III datasets.

\section{Conclusion and Future Work} 
\label{sec:conclusion}

We proposed a novel model, the HorseshoeBNN, for performing interpretable patient outcome prediction. Our method extends traditional BNNs to perform feature selection using sparsity inducing prior distributions in a tied manner. Our architecture offers many advantages. Firstly, being based on a BNN, it represents a non-linear, fully probabilistic method which is highly compatible with the clinical decision making process. Secondly, with our proposed advances, the model is able to learn which input features are important for prediction, thereby making it interpretable, which is highly desirable in the clinical domain. 

We worked closely with clinicians and evaluated our model using two real-world ICU cohorts. We showed that our proposed HorseshoeBNN can provide additional insights about the importance of input features. Together with its ability to provide uncertainty estimates, the HorseshoeBNN could be used to support clinicians in their decision making process.
In view of the high-dimensional complex nature of medical data and the high relevance of outcome prediction in healthcare, our method could be useful not only for ICUs but in any medical settings. Our work illustrates how a close collaboration between computational and clinical experts can lead to methodological advances suitable for translation into tools for patient benefit.

In future work, we will extend our model to be able to work with the entire time series \cite{yoon2018deep} as dynamic prediction is a key area of interest in the ICU.
Utilizing information about the evolution of features over time might not only improve predictive accuracy, but could provide additional insights about temporal changes of measurement values. 

Moreover, we will use more sophisticated methods for missing value imputation to obtain better predictive performance \cite{ma2018eddi,little2019statistical}. 
Finally, contemporary digital healthcare is fundamentally transdisciplinary and we would like to continue working with medical experts to explore how to deploy our method in a real-world clinical setting.

\section*{Acknowledgements}
CENTER-TBI is supported by The European Union FP 7th Framework program (grant 602150) with additional funding provided by the Hannelore Kohl Foundation (Germany) and by the non-profit organization One Mind For Research (directly to INCF).

{\fontsize{9.0pt}{10.0pt}
\selectfont
\bibliography{bibliography.bib}
\bibliographystyle{aaai}}

\clearpage

{\appendix\setcounter{secnumdepth}{2}}
\onecolumn
\section{Validation of Induced Sparsity}
\label{app:sparsity}

In this section, we verify the sparsity-inducing capacities of our LinearHorseshoe model. We repeat an experiment proposed in \cite{hernandez2015a}. In the experiment, a data matrix $X$ with 75 datapoints is sampled from the unit hypersphere. Target values $y$ are computed using $y = X \cdot w + \epsilon$, where $\epsilon$ represents Gaussian distributed noise with standard deviation 0.005. The 512-dimensional weight vector $w$ is sparse, that is, only 20 randomly selected components are non-zero. 

Because the target values $y$ depend linearly on the data matrix $X$, we use a linear model with a horseshoe prior to obtain an estimate $\tilde{w}$ of the weight vector. The estimate $\tilde{w}$ is given by the mean of the posterior distribution. We evaluate the model using the reconstruction error $\|\tilde{w}-w\| / \|w\|$ and find results similar to those reported in \cite{hernandez2015a}. Hyperparameter settings can be found in Table \ref{table:params-prior-experiment}. The results of this validation experiment show that our model is capable of correctly reproducing sparse weight vectors.\\

\begin{table}[h!]
\centering
\begin{tabular}{ l  l}
\toprule
  \bf{Hyperparameter} & \bf{Value} \\
  \midrule
  Number of weight samples during training & 10 \\
  Number of weight samples during testing & 100 \\
  Batch size & 64 \\
  Learning rate & 0.001 \\
  Number of epochs & 2000000 \\ 
  Global shrinkage parameter $b_g$ of Horseshoe prior & 1.0\\
  Local shrinkage parameter $b_0$ of Horseshoe prior & 1.0\\
  \bottomrule
\end{tabular}
\caption{Parameter and hyperparameter settings for prior experiment on induced sparsity}
\label{table:params-prior-experiment}
\end{table}

\begin{figure}[h!]
    \centering
    \includegraphics[width = 0.5 \textwidth]{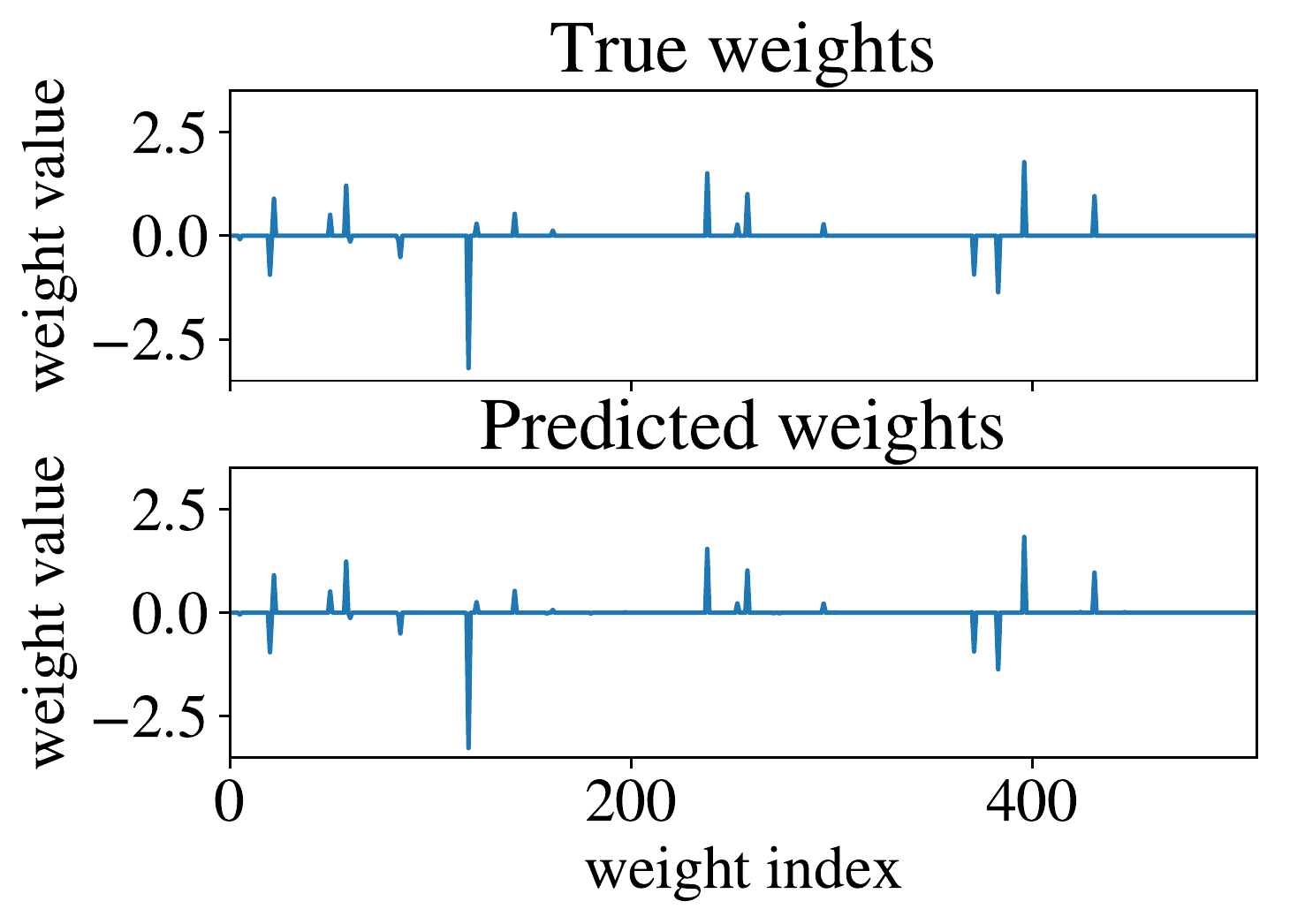}
    \caption{Signal reconstruction (bottom) of a sparse signal (top).}
    \label{fig:reconstruction}
\end{figure}

The average reconstruction error over twenty realizations of the experiment is $0.399 ~\pm~0.272$. This error is slightly higher but of the same order of magnitude as the error of $0.16 ~\pm~0.07$ reported in \cite{hernandez2015a}. The difference can be explained by the fact that we use variational inference to approximate the posterior distribution whereas \cite{hernandez2015a} use Markov chain Monte Carlo techniques, which are known to give a more accurate approximation of the posterior distribution. An example of a sparse signal and the reconstructed weights is shown in Figure~\ref{fig:reconstruction}.

\section{Models}
\label{app:models}
For the comparison with existing models, we make use of feature selection models implemented in \textit{Scikit Learn}. For the benchmarking on regression datasets from the UCI repository, we compare to the \textit{Lasso} model, the \textit{SVR} model (a support vector regressor) with linear kernel and the \textit{RandomForestRegressor}. For classification tasks we use the \textit{LogisticRegression} model with  $\ell_1$ regularization (equivalent to Lasso), the \textit{LinearSVC} model with  $\ell_1$ regularization and the \textit{RandomForestClassifier}). For all models the regularization strength is optimized to minimize the test error.
\clearpage
\section{MIMIC-III}
\label{app:mimic}
\subsection{Feature ranges and hyperparameters}
\label{app:mimic_params}
Allowed ranges for the features in the MIMIC-III dataset are shown in Table \ref{tab:mimic}. Table \ref{tab:params-mimic} shows the hyperparameters for the horseshoeBNN. 

\begin{table*}[h!]
    \small
\begin{center}
    \begin{tabular}{  l  l  p{1.7cm}  p{1.7cm} }
    \toprule
    \textbf{Feature} &
    \textbf{Unit} &
    \textbf{Min threshold} & \textbf{Max  threshold}  \\
    \midrule
    Height & cm & $>0$ & 250\\
    Temperature & \textdegree C & 20 & 49 \\
    Blood pH$^*$ & - & 6 & 8 \\
    Fraction of inspired oxygen & - & 0 & 1 \\
    Capillary refill time & seconds & 0 & - \\
    Heart rate & bpm & 0 & 300 \\
    Systolic blood pressure & mmHg & $>0$ & 275 \\
    Diastolic blood pressure & mmHg & $>0$ & 150 \\ 
    Mean blood pressure & mmHg & $>0$ & 190 \\
    Weight & kg & 0 & 250 \\
    Glucose & mg/dL & 0 & 1250 \\ 
    Respiratory rate & number breaths per min & 0 & 150 \\
    Oxygen saturation & \% & $>0$ & 100 \\
    Glasgow coma scale eye response & - & 1 & 4 \\
    Glasgow coma scale motor response & - & 1 & 6 \\
    Glasgow coma scale verbal response & - & 1 & 5 \\
    Glasgow coma scale total & - & 3 & 15\\
    \bottomrule
    \end{tabular}
\end{center}
\caption{MIMIC-III feature ranges $^*${\footnotesize For blood pH values higher than 14 (physically impossible), we assume that these are actually measures of hydrogen ion concentrations in nanomole/L. These values are converted to pH values, after which the threshold is applied.}}
\label{tab:mimic}
\end{table*}

\begin{table}[h!]
\centering
\begin{tabular}{l  l}
\toprule
  \bf{Hyperparameter} & \bf{Value} \\
  \midrule
  Number of weight samples during training & 10 \\
  Number of weight samples during testing & 100 \\
  Batch size & 64 \\
  Number of hidden units & 50 \\
  Learning rate & 0.001 \\
  Number of epochs & 5000 \\
Standard deviation of Gaussian prior & 1.0\\
  Global shrinkage parameter $b_g$ of Horseshoe prior & 1.0\\
  Local shrinkage parameter $b_0$ of Horseshoe prior & 1.0\\
  \bottomrule
\end{tabular}
\caption{Parameter and hyperparameter settings for MIMIC models}
\label{tab:params-mimic}
\end{table}

\subsection{Confusion matrices}

\begin{figure}[h!]
    \centering
    \includegraphics[width = 0.9 \textwidth]{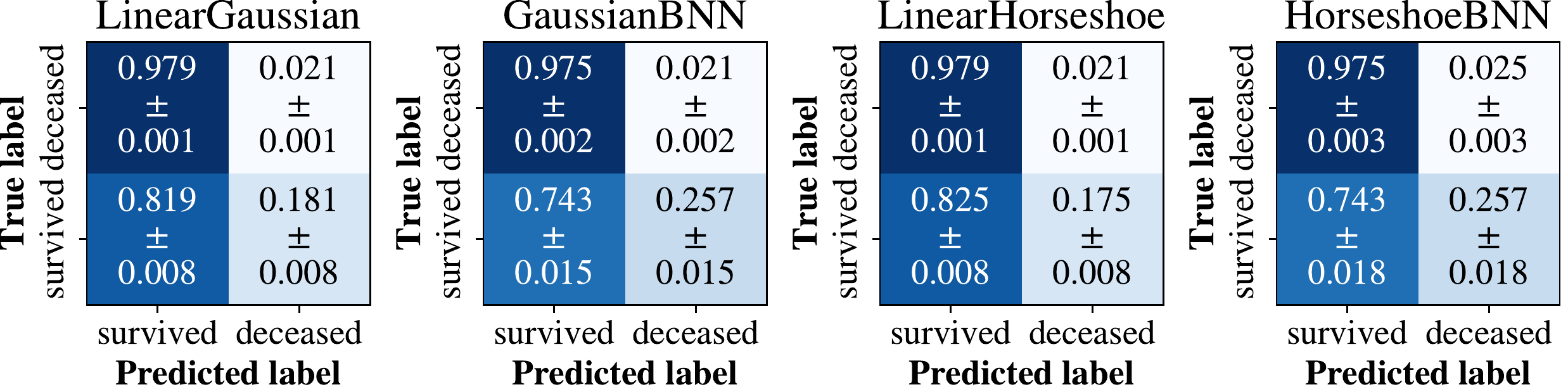}
    \caption{Results of the different models for the task of mortality prediction tested on the MIMIC-III cohort. The mean value and one standard error of each metric over 10-fold cross-validation is presented.}
    \label{fig:mimic_confusion_all}
\end{figure}

Figure \ref{fig:mimic_confusion_all} shows the confusion matrices for the models trained on the MIMIC-III dataset.

\subsection{Weight histograms}
\label{app:histogram}
\begin{figure}[h!]
    \centering
    \includegraphics[width = 0.9 \textwidth]{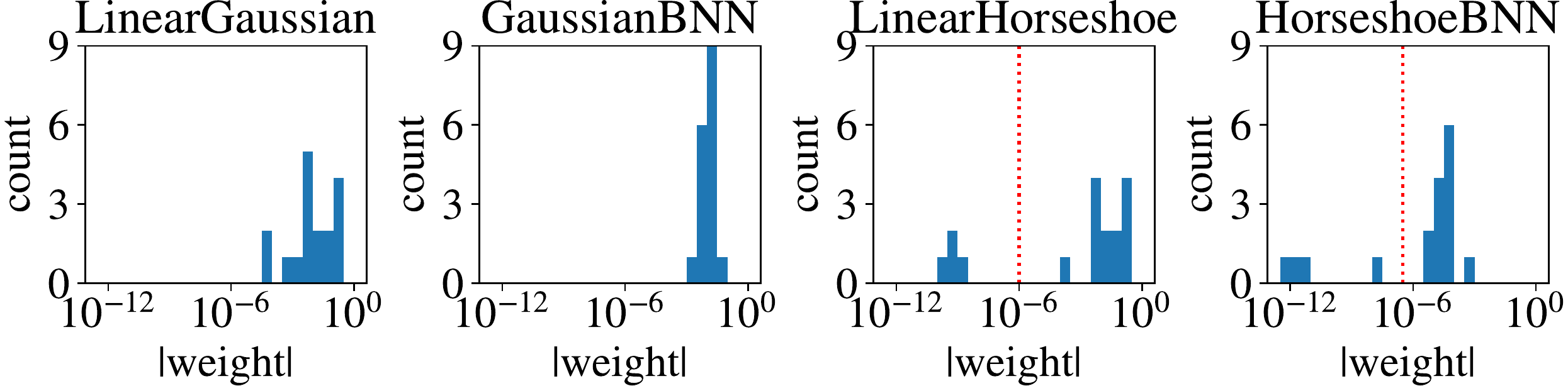}
    \caption{Histograms of the weights of the first layers of the models. Horseshoe models show more significant shrinkage of irrelevant weights.}
    \label{fig:hist_mimic}
\end{figure}

We inspect the histograms of the means of the distributions of the weights of the LinearHorseshoe and LinearGaussian models. For the GaussianBNN and the HorseshoeBNN, we calculate the average weight per feature in the first layer. These weights are shown in the histograms in Fig. \ref{fig:hist_mimic} The histograms corresponding to the LinearHorseshoe model and the HorseshoeBNN models show a clearer separation into two groups (see red dotted lines): the irrelevant weights shrink more because of the horsehoe prior. This demonstrates that models with a horseshoe prior are better suited for feature selection than models with Gaussian priors.

\subsection{Alternative threshold determination}
\label{app:threshold}

\begin{figure}[h!]
    \centering
    \includegraphics[width = 0.9 \textwidth]{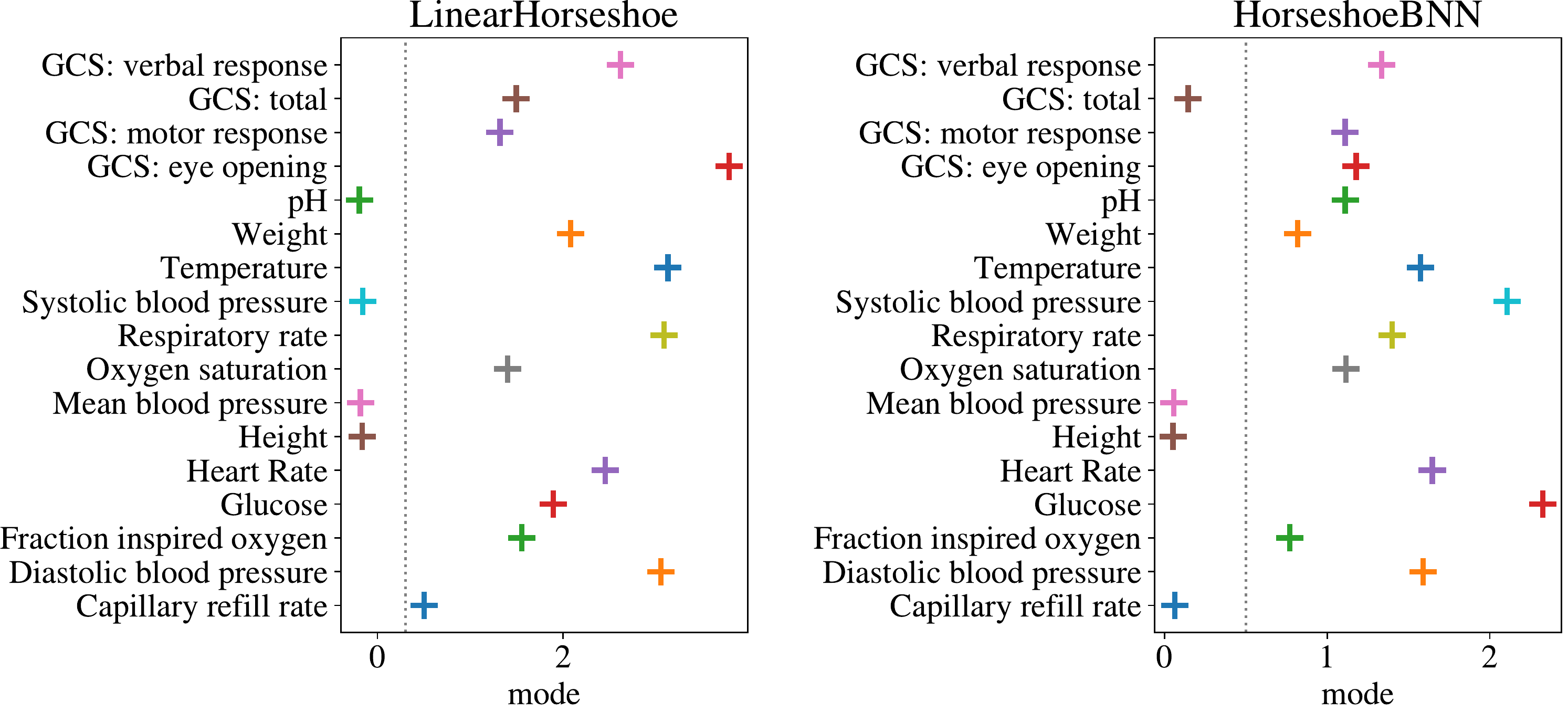}
    \caption{Mode of the posterior distribution of $v\tau_j$ for all features $j$ as determined by the LinearHorseshoe model (left) and the HorseshoeBNN (right). The dotted line indicates the threshold for feature relevance.}
    \label{fig:modes}
\end{figure}

The importance of the $j$th feature can also be determined by inspecting the posterior distribution of the scale parameters $v$ and $\tau_j$. As can be seen in Equation \ref{eq:logn}, the product $v\tau_j$ follows a log-normal distribution. \citet{louizos2017} determine relevance by setting a threshold for the mode of this distribution. The modes of the posterior distributions of $v\tau_j$ for the features as determined by the LinearHorseshoe model are shown in Fig. ~\ref{fig:modes}. The features left of the dotted threshold line (\textit{Height}, \textit{Mean blood pressure}, \textit{Systolic blood pressure} and \textit{pH}) are considered irrelevant by the model. This is in agreement with the findings in Fig.~\ref{fig:importance_mimic} in the main text. Figure \ref{fig:modes} shows a similar result for the mode of the posterior distribution of $v\tau_j$ for the HorseshoeBNN. Again, the features considered irrelevant are the same as in Fig.~\ref{fig:importance_mimic}.

\clearpage
\section{CENTER-TBI}
\label{app:tbi}

\subsection{Feature ranges and hyperparameters}
\label{app:tbi_params}
The features in the CENTER-TBI dataset are listed in Table \ref{table:TBI}. Table \ref{table:params-tbi} shows the hyperparameters for the horseshoeBNN.

\begin{table*}[h!]
\centering
\begin{tabular}{l  l}
  \toprule
  \bf{Category} & \bf{Features} \\
  \midrule
  General & Age, gender, prior alcohol use, history of anti-coagulants\\
  Injury& Cause of injury, time of injury \\
  Condition on arrival & heart rate, respiratory rate, temperature, SpO$_2$\\
  & Systolic blood pressure, diastolic blood pressure \\
  & Arterial O$_2$ tension, CO$_2$ tension, pH \\
  & Assessment of airway, breathing, circulation\\
  & Episode of hypoxia or hypotension\\
  Neurological assessment & Glasgow Coma Score, pupil reaction\\
  Initial imaging & Marshall classification, depressed skull fracture\\
  & subarachnoid hemorrhage, midline shift\\
  & Absent basal cisterns, extradural hematoma\\
  Blood chemistry tests & Glucose, sodium, albumin, calcium, hemoglobin\\
  & hematocrit, white blood cell count, C-reactive protein,\\
  &  Platelet count, International normalized ratio\\
  & activated partial thromboplastin time, fibrogen\\
  \bottomrule
\end{tabular}
\caption{Features used to predict mortality on the CENTER-TBI dataset}
\label{table:TBI}
\end{table*}

\begin{table}[h!]
\centering
\begin{tabular}{l  l}
\toprule
  \bf{Hyperparameter} & \bf{Value} \\
  \midrule
  Number of weight samples during training & 10 \\
  Number of weight samples during testing & 100 \\
  Batch size & 64 \\
  Number of hidden units & 100 \\
  Learning rate & 0.001 \\
  Number of epochs & 5000 \\
  Standard deviation of Gaussian prior & 1.0\\
  Global shrinkage parameter $b_g$ of Horseshoe prior & 1.0\\
  Local shrinkage parameter $b_0$ of Horseshoe prior & 1.0\\
  \bottomrule
\end{tabular}
\caption{Parameter and hyperparameter settings for CENTER-TBI models}
\label{table:params-tbi}
\end{table}

\subsection{Confusion matrices}

\begin{figure}[h!]
    \centering
    \includegraphics[width = 0.9 \textwidth]{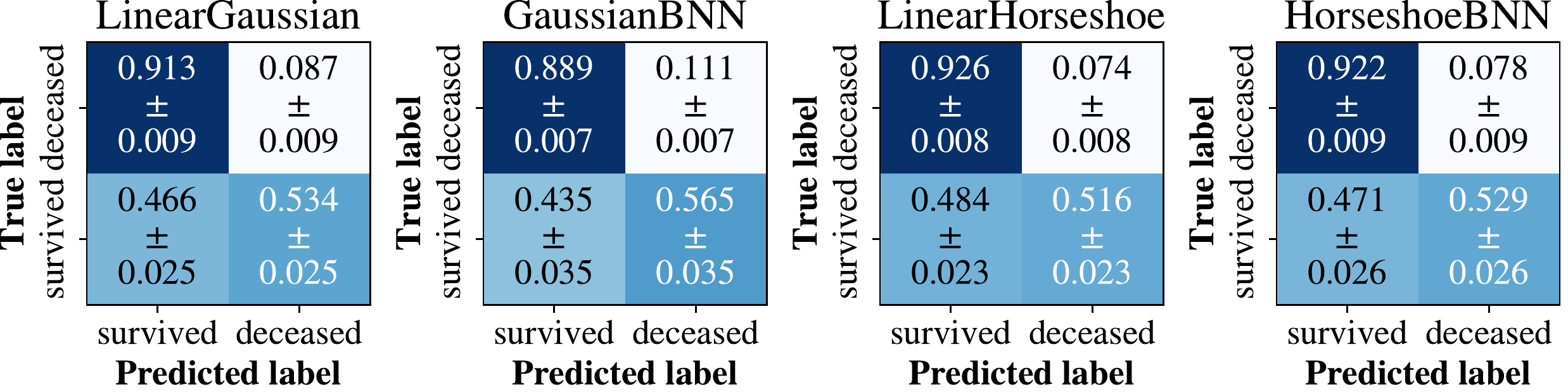}
    \caption{Results of the different models for the task of mortality prediction tested on the CENTER-TBI cohort. The mean value and one standard error of each metric over 10-fold cross-validation is presented.}
    \label{fig:tbi_confusion}
\end{figure}

Figure \ref{fig:tbi_confusion} shows the confusion matrices for the models trained on the CENTER-TBI dataset. Just as for the MIMIC-III dataset, the BNN based models are better at predicting the outcome for deceased patients than the linear models.

\subsection{Feature importance}
\begin{figure*}[h!]
    \centering
    \includegraphics[width = 0.9\textwidth]{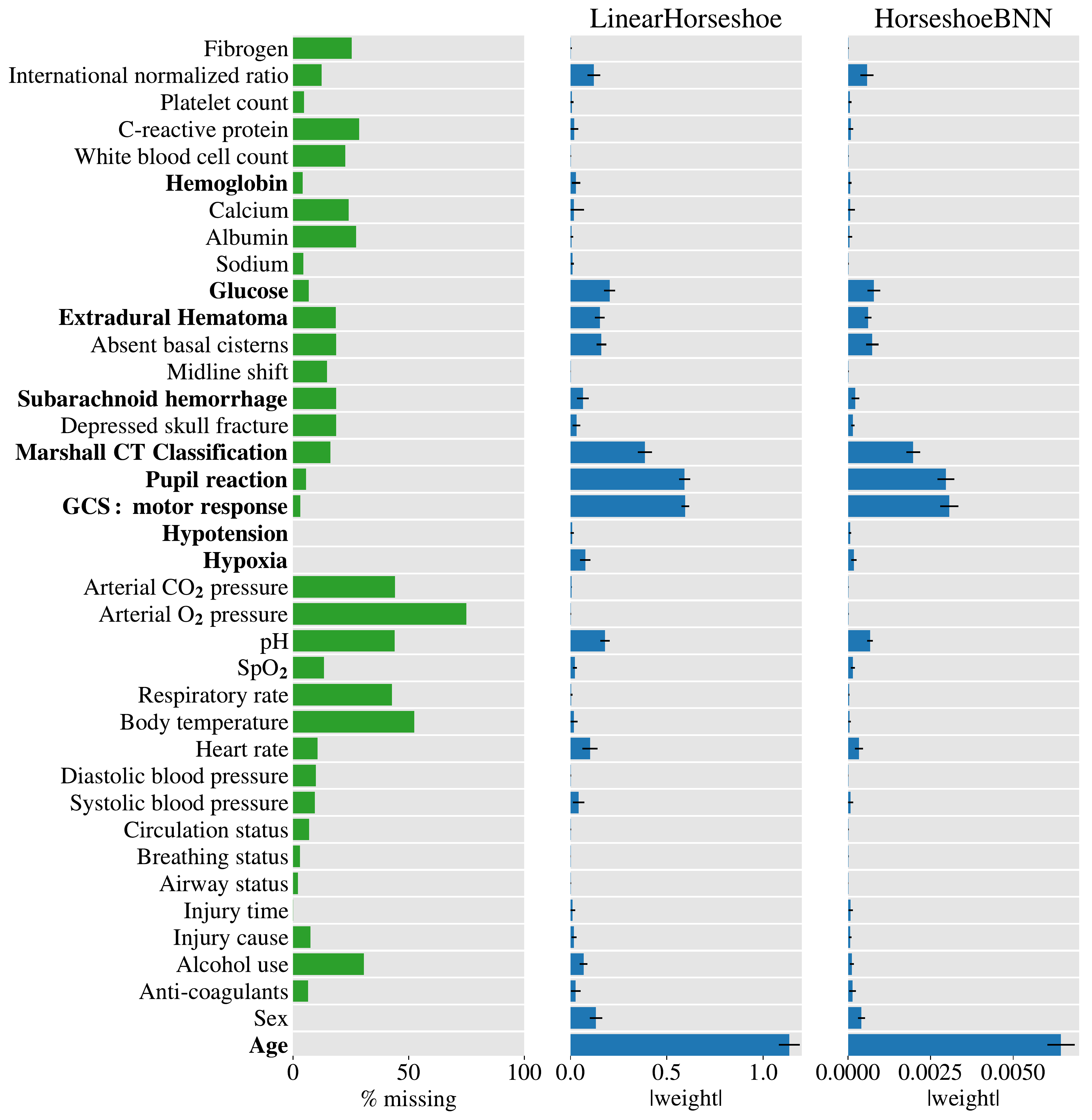}
    \caption{\textbf{Left:} Percentage of missing data for the features in the CENTER-TBI cohort. \textbf{Middle:} Norm of the weights of the LinearHorseshoe model, representing the relative importance of the corresponding features. \textbf{Right:} Norm of the weights of the HorseshoeBNN. The name of each input feature is given on the far left of the plots. Features included in the IMPACT model for outcome prediction in TBI are marked bold. Feature weights of zero indicate that the corresponding features are irrelevant for outcome prediction. All non-zero weights indicate that the corresponding features are relevant for predicting mortality.}
    \label{fig:importance_tbi}
\end{figure*}

The full graph of the relevance of the different input features as determined by the LinearHorseshoe model and the HorseshoeBNN are shown in Figure \ref{fig:importance_tbi}. Features marked in bold are included in the IMPACT model \cite{steyerberg2008}, a model commonly used for outcome prediction in TBI.

\subsection{Feature importance - RandomForest}

\begin{figure*}[h!]
    \centering
    \includegraphics[width = 0.7\textwidth]{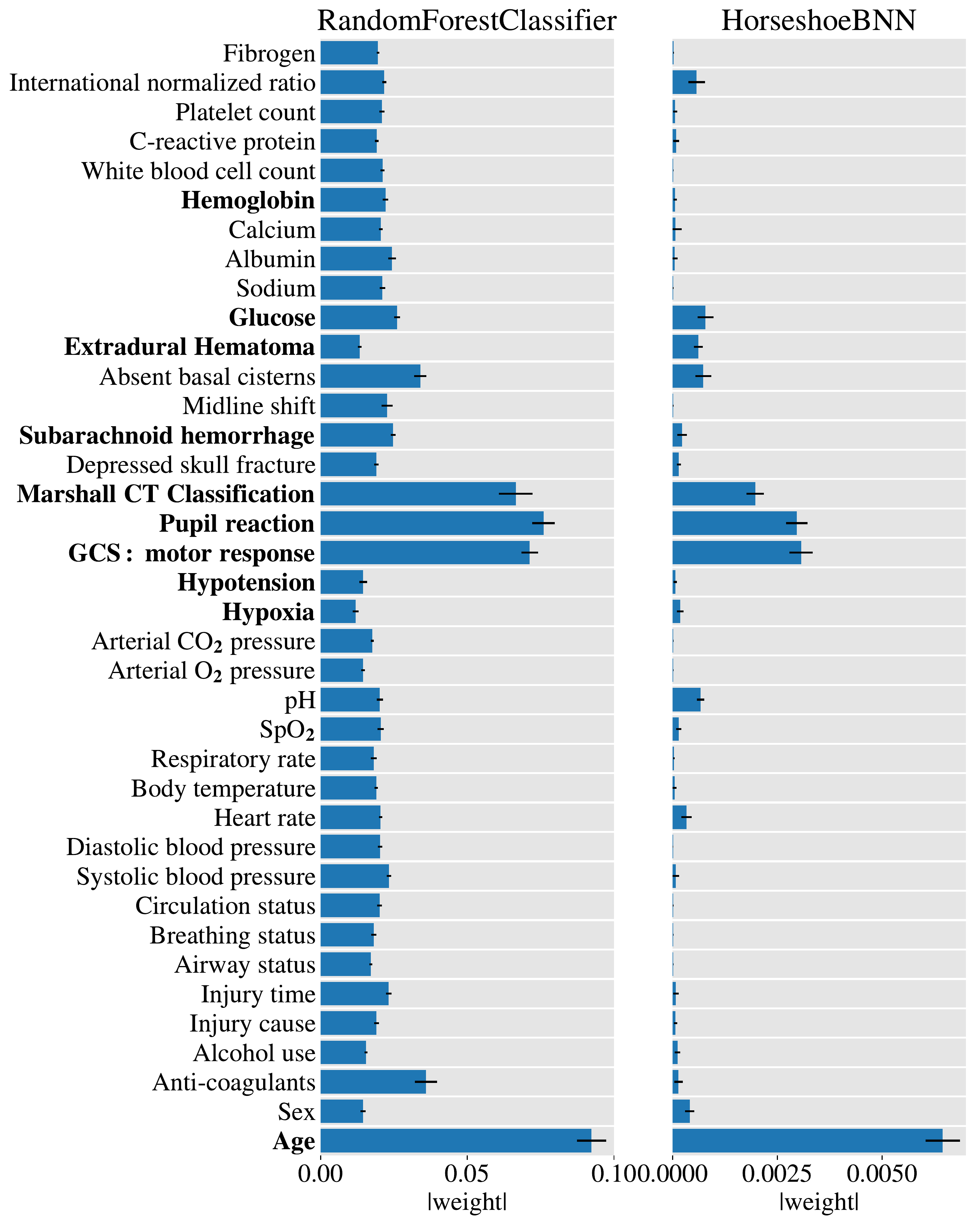}
    \caption{\textbf{Left:} Norm of the weights of the RandomForest \textbf{Right:} Norm of the weights of the HorseshoeBNN.}
    \label{fig:importance_forest}
\end{figure*}

Figure \ref{fig:importance_forest} compares the relevance of features attributed by the RandomForest and the HorseshoeBNN. The models agree on the four most important features, but for the RandomForest all other weights are roughly equal in size. It is therefore unable to distinguish other relevant features.
\clearpage
\section{Discussion of feature selection results}
\label{app:feature-selection-results}
\begin{figure*}[h!]
    \centering
    \includegraphics[width = 1 \textwidth]{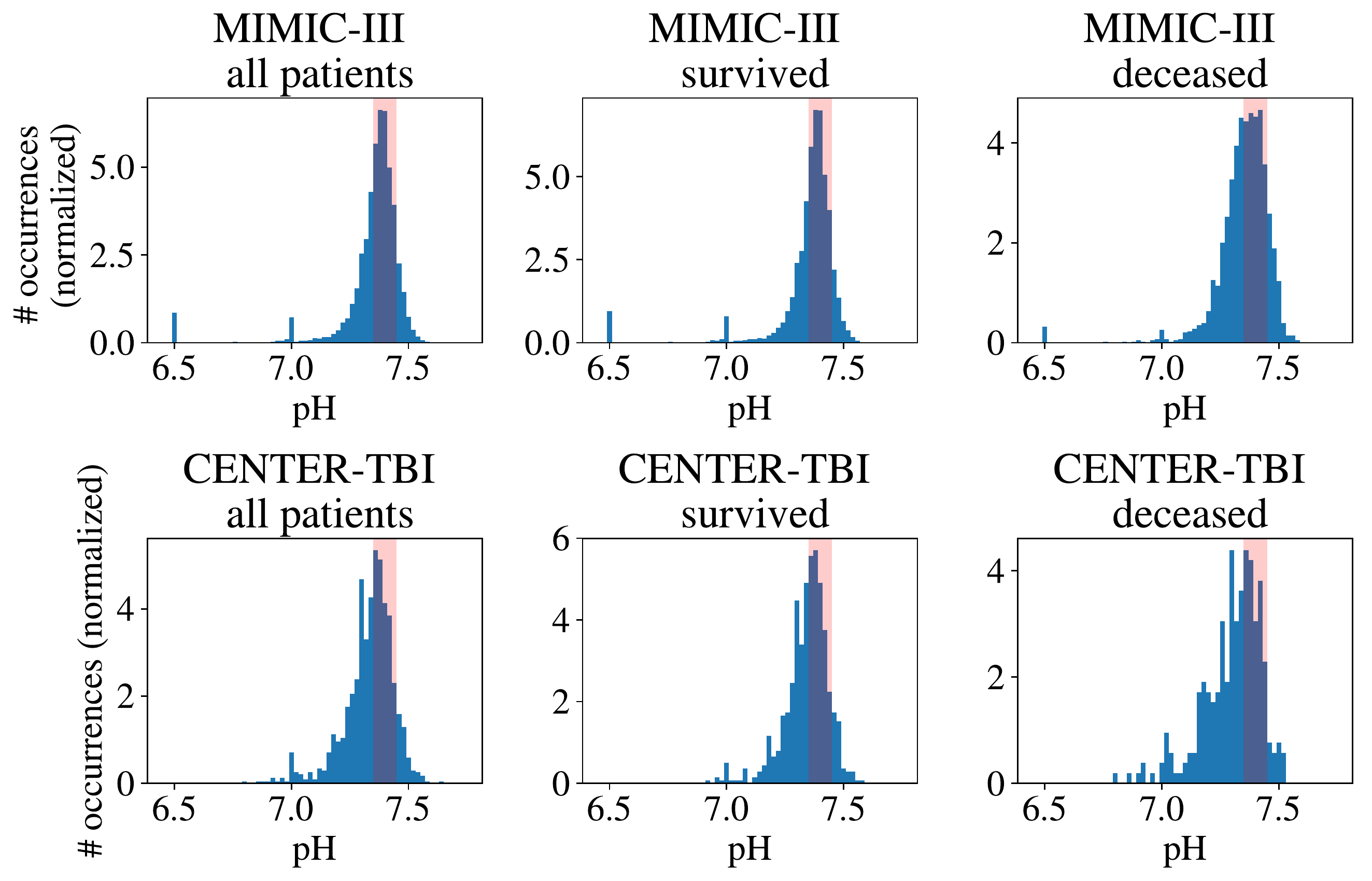}
    \caption{Distribution of the feature \textit{pH} in the MIMIC and CENTER-TBI cohorts. Illustrated in red is the range of pH for healthy patients (7.35-7.45).}
    \label{fig:pH}
\end{figure*}

When comparing with the results for MIMIC-III we observe different results for the feature \textit{pH}. Whereas both horseshoe models trained on CENTER-TBI determine pH to be important, this is not the case for MIMIC-III. For the latter only the non-linear model, that is, the HorseshoeBNN, selects this feature. This may be due to a difference between the patient groups contained in MIMIC-III and CENTER-TBI. This assumption is further supported by the distribution of pH values in the cohorts. This can be explained by the difference between the patient groups contained in MIMIC vs. CENTER-TBI. Whereas MIMIC contains a very heterogeneous group of patients, CENTER-TBI contains only patients that experienced some kind of trauma , and the outcome predictors will therefore be expected to be very different given this is a different disease. Figure \ref{fig:pH} illustrates the distribution of pH in both datasets. When computing the KL-divergence between the distributions for patients that survived and patients that deceased we observe that the divergence is larger for CENTER-TBI. This might explain why both models determine pH to be important for CENTER-TBI, whereas only the non-linear model includes pH for MIMIC.

\begin{figure*}[h!]
    \centering
    \includegraphics[width = 1 \textwidth]{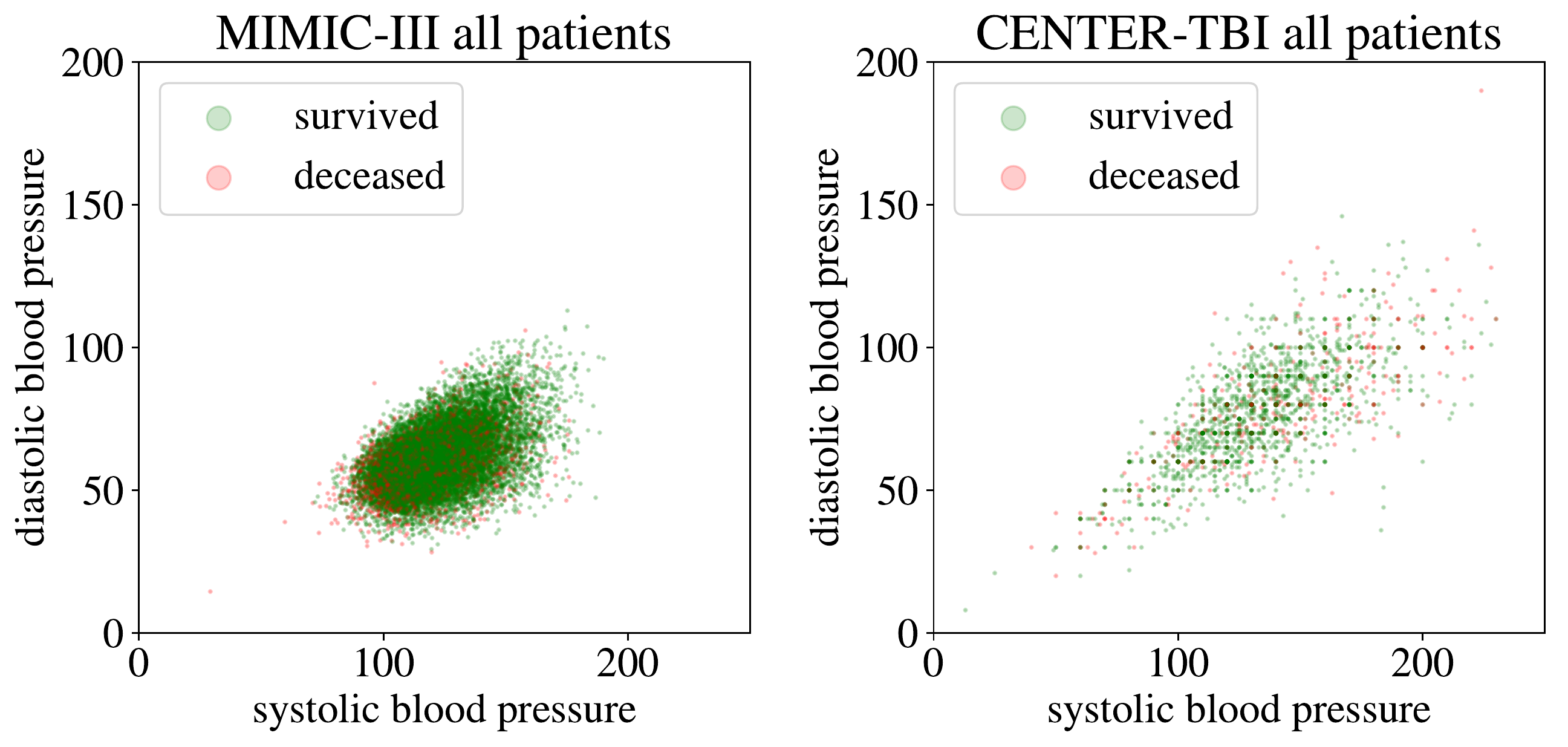}
    \caption{Correlations between the features \textit{diastolic blood pressure} and \textit{systolic blood pressure} in the MIMIC and CENTER-TBI cohorts.}
    \label{fig:blood_pressure}
\end{figure*}

A further difference was observed for blood pressure. Both CENTER-TBI horseshoe models indicate that only systolic blood pressure is of relevance, whereas for MIMIC-III the HorseshoeBNN determined both systolic and diastolic blood pressure to be important. Again, this could be explained by the heterogeneity of the datasets. CENTER-TBI contains predominantly people with relatively isolated brain injuries not affecting the blood circulation. Therefore, blood pressure (absolute, e.g. systolic) is most important as this pressure perfuses the brain. In contrast, MIMIC-III includes a large number of patients in shock (e.g. with sepsis). 
The degree of this pathology, which would be less likely to occur in the CENTER-TBI patients, is clinically likely to be a strong determinant of outcome and would be expected to be reflected in a difference in diastolic blood pressure. This is further supported when inspecting the relationship between systolic and diastolic blood pressure in both datasets. Figure \ref{fig:blood_pressure} shows that the two features are strongly correlated for CENTER-TBI but not for MIMIC. This might explain why only the systolic blood pressure is considered relevant for CENTER-TBI, but both the systolic and diastolic blood pressure for MIMIC.

\end{document}